\documentclass[preprint,12pt]{elsarticle}

\usepackage{setspace} 
\usepackage{amssymb}
\usepackage{graphicx}
\usepackage{color}
\usepackage{booktabs}
\usepackage{mathrsfs}
\usepackage{multirow}
\usepackage{amsmath}
\usepackage[ruled]{algorithm2e}
\newtheorem{theorem}{Theorem}
\newtheorem{proof}{Proof}[section]
\usepackage[caption=false,font=normalsize,labelfont=sf,textfont=sf]{subfig}

\journal{Pattern Recognition}

\begin{document}
	
	\begin{frontmatter}

		\title{Learning Cross-domain Semantic-Visual Relationships for Transductive Zero-Shot Learning}
		\author[label1]{Fengmao Lv\corref{cor5}}
		\author[label2]{Jianyang Zhang\corref{cor5}}
		\author[label2]{Guowu Yang}
		\author[label3]{Lei Feng}
		\author[label4]{Yufeng Yu}
		\author[label2]{Lixin Duan}
		\cortext[cor5]{Corresponding authors. \\Email addresses: fengmaolv@126.com, jianyangzhang@std.uestc.edu.cn}
		\address[label1]{Southwest Jiaotong University, West Park of High-tech Zone, Chengdu, 611756, Sichuan, China}
		
		\address[label2]{University of Electronic Science and Technology of China, No.2006, Xiyuan Ave, Chengdu, 611731, Sichuan, China}
		
		\address[label3]{Chongqing University, No.174 Shazhengjie, Chongqing, 400044, China}
		\address[label4]{Guangzhou University, No.230 Wai Huan Xi Road, Guangzhou, 510006, Guangdong, China}
		\begin{abstract}

			Zero-Shot Learning (ZSL) learns models for recognizing new classes.  One of the main challenges in ZSL is the  domain discrepancy  caused by the category inconsistency  between training and testing data. Domain adaptation  is the most intuitive way to address this challenge. However, existing domain adaptation techniques cannot be directly applied into ZSL due to the disjoint label space between source and target domains. This work proposes the Transferrable Semantic-Visual Relation (TSVR) approach towards transductive ZSL. TSVR redefines image recognition as predicting the similarity/dissimilarity labels for semantic-visual fusions consisting of class attributes and visual features. After the above transformation, the source and target domains can have the same label space, which hence enables to quantify domain discrepancy. For the redefined problem, the number of similar semantic-visual pairs is significantly smaller than that of dissimilar ones. To this end, we further propose to use Domain-Specific Batch Normalization to align the domain discrepancy.
		\end{abstract}

		\begin{keyword}
			Zero-shot learning \sep transfer learning \sep domain adaptation.
		\end{keyword}
		
	\end{frontmatter}

	\section{Introduction}
	Although deep learning has achieved advances in image recognition over the past years, the corresponding performance leaps heavily rely on collecting sufficient labeled data for each category to be recognized \cite{xian2018zero}. Due to the exponential growth of image data and potential classes, it usually requires a huge amount of time and human labor to collect well-labeled training data for new classes. This severely prevents deep learning from generalizing its prediction ability to new classes. To tackle this issue,  Zero-Shot Learning (ZSL), which is originally inspired by humans' ability to recognize new objects without seeing samples, has attracted increasing attentions recently.
	
	ZSL aims to learn recognition models which transfer knowledge from seen (source) classes with labeled samples to recognize novel (target) classes without labeled samples \cite{cao2023review}. The main idea towards ZSL is to bridge the source and target classes by an intermediate-level semantic representation, which can be defined by attributes \cite{farhadi2009describing, wah2011caltech}, or word2vec. Semantic representations which are assumed to be shared between the source and target classes constitute the key factor for cross-category transfer. To this end, most existing ZSL methods project visual features into the semantic representations and recognize images from novel target categories by the nearest neighbor search in the shared space \cite{akata2013label,kodirov2017semantic}. In practice, the performance of ZSL is usually effected by the domain shift problem \cite{fu2015transductive}. However, the domain shift problem studied in the existing ZSL methods~\cite{guo2016transductive,xie2021further} differs from the conventional domain shift problems. In particular, it is indirectly observed in terms of the projection shift rather than the feature distribution shift across the training and testing data. This kind of domain shift problem is also usually called as projection domain shift or project shift~\cite{guo2016transductive}. On the other hand, the conventional domain shift problem is mainly studied in the unsupervised domain adaptation task which sets the identical label space for the source and target domains~\cite{long2015learning}. For the ZSL problem, as the source and target domains have different label spaces, we cannot directly define the distribution discrepancy for the ZSL problem. Currently, it is still unknown how the distribution shift alignment directly effects the performance of ZSL.

	Motivated by the above observation, we propose a novel approach dubbed Transferrable Semantic-Visual Network (TSVN) by drawing on an intriguing insight connecting two transfer paradigms, i.e., domain adaptation and zero-shot learning.  Our approach mainly focuses on investigating the effect of the direct distribution shift on the performance of ZSL by transforming ZSL into a domain adaptation problem. To this end, we redefine the image recognition problem as predicting the similarity/dissimilarity labels for the semantic-visual fusions consisting of class attributes and visual features. The source and target domains can share the identical classes (i.e., the similarity and dissimilarity classes) after the redefinition, although the original ZSL problem sets disjoint label space for them. For this redrawn domain adaptation problem, the distribution shift of semantic-visual fusions across the source and target domains will establish the performance bottleneck. Hence, our approach is performed under the transductive setting and introduces the unlabeled target domain data to reduce the distribution shift as in the standard unsupervised domain adaptation problem.
	
	To this end,  a crucial point lies in how to conduct distribution alignment for the representations of semantic-visual fusions  across the source and target domains. In general, we can use MMD \cite{long2015learning} or domain adversarial training \cite{tzeng2017adversarial} to align their distributions within each mini-batch. However, for the redrawn domain adaptation problem to be tackled in this work, the class distribution is extremely imbalanced, {i.e.}, the number of the similar semantic-visual pairs is significantly smaller than that of the dissimilar ones. For example, if the prediction task involves $K$ categories, the number of the negative samples will be $K-1$ times as large as that of the positive samples. Hence, there can be very few positive samples in each mini-batch. As a result, distribution alignment will be dominated by the large category if it is conducted over mini-batches. In this work, we propose to use a mild but effective component dubbed Domain-Specific Batch Normalization (DSBN) to align the distribution of semantic-visual fusions. Specifically, we incorporate two Batch Normalization (BN) units at each layer. This design enables the neural network to separately normalize the mini-batches from different domains with {zero} mean and {one} variance. Hence, the semantic-visual fusions from the source and target domains can have similar distribution at each layer of the neural network. Although the number of similar semantic-visual fusions is very small in each mini-batch, the moving average mechanism of batch normalization enables to retain their effect in feature alignment. Our approach is similar to Adaptive Batch Normalization (ABN) proposed from \cite{li2018adaptive}.  However, ABN only maintains the normalization statistics of source data at training and requires a separate process to modulate the normalization statistics for the target domain. Note that the target data cannot be used for training the network in the ABN approach since they will cause bias in calculating the statistics of the source domain. In contrast, our  design allows us to train the network by the target data through incorporating extra practical  regularization terms over the target samples.

	Overall, the main contribution of this work is two-fold: 1) We propose to investigate the effect of the direct  distribution shift on the performance of zero-shot learning by transforming ZSL into a standard domain adaptation problem.  2) We introduce the domain-specific batch normalization component to reduce the distribution mismatch for the representations of semantic-visual fusions across the source and target domains. Compared with the previous ZSL works which can only indirectly bridge the projection shift~\cite{xian2018feature}, our approach provides a more direct path to bridge the distribution shift across the source and target domains. Experimental results over diverse ZSL benchmarks clearly demonstrate the superiority of our method compared with the existing ZSL methods.
	
	The rest of this paper is organized as follows.  Section~\ref{related_work} reviews the related works on zero-shot learning and domain adaptation.  Section~\ref{approach} introduces the theoretical insights and technical details of the TSVR approach. Section~\ref{expirements} reports  experimental results on different benchmarks to validate the effectiveness of TSVR. Finally, Section~\ref{conclusion} concludes this paper.
	
	\section{Related Work}
	\label{related_work}
	\subsection{Zero-shot learning}
	Zero-Shot Learning aims to learn recognition models for recognizing new classes. The main strategy for ZSL is to  associate source and target classes through an intermediate-level semantic representation. The semantic space can be defined by attributes \cite{farhadi2009describing, wah2011caltech}, text description, or word2vec.  In general, most existing ZSL methods project visual features into the semantic space and recognize images from novel target categories by the nearest neighbor search in the semantic space \cite{akata2013label,kodirov2017semantic}. Unfortunately, the nearest neighbor search in semantic space will suffer from the problem of hubness surfaces \cite{fu2015transductive}. To tackle this limitation, the recent works propose to project semantic attributes into visual space as prototypes \cite{xian2018feature,sung2018learning} or use semantic attributes to refine visual features \cite{chen2021free, cao2022mff}. In addition,  the visual features and the semantic representations can also be associated by projecting them into a shared intermediate space \cite{chen2021hsva}. However, the projection function learned from the source domain tends to be biased when they are directly applied to target images, which is usually named as domain shift or projection shift \cite{fu2015transductive}.
	
	\subsection{Transductive zero-shot learning}
	Transductive ZSL aims to relieve the projection shift between the source and target categories through introducing  unlabeled images from novel classes during training \cite{fu2015transductive}. To be specific, Kodirov {et al.} \cite{kodirov2015unsupervised} propose to adapt the projection function by regularized sparse coding. Akata {et al.} \cite{akata2015label} propose label propagation to refine the results for unlabeled target samples. Guo {et al.} \cite{guo2016transductive} propose  a  joint leaning approach considering both source and target classes
	simultaneously to learn the shared model space.
	Later, Fu {et al.} \cite{fu2015transductive} propose a multi-view semantic space alignment process to alleviate the projection shift with multiple semantic views. More recently, Li {et al.} \cite{li2019leveraging} propose to leverage the target images with high confidence as the references to recognize other target data.
	
	\subsection{Domain adaptation}
	Domain adaptation aims at transferring knowledge across two different domains that share the identical task/label space. In general, the major challenge in domain adaptation lies in the distribution discrepancy between the source and the target domains. Therefore, the natural idea towards domain adaptation is to learn deep representations that can align the distribution shift between two domains. In general, the existing domain adaptation methods mainly focus on reducing the distribution discrepancy~\cite{long2015learning, zellinger2019robust, shen2018wasserstein} or performing domain adversarial training \cite{tzeng2017adversarial} at intermediate layers of deep neural networks. In addition to the above methods, Li {et al.} \cite{li2018adaptive} propose that the statistics of BN layers is curtail for aligning the distribution mismatch. Compared with \cite{li2018adaptive}, our design enables to train the network with both the source and target data. Hence, we can incorporate extra practical regularization terms for the target data during the training phase.

	\section{Transferrable Semantic-Visual Relationship}
	\label{approach}

	\subsection{Problem Statement} 
	Formally, the task of ZSL aims to learn recognition models for recognizing new classes without labeled data. For the source domain, we are given a set of source categories $\mathcal{C}^s=\{c_{1}^{s},...,c_{K_s}^{s} \}$ and a labeled dataset of source images $\mathcal{D}^s =\{(\textbf{x}_1^{s},\textbf{y}_1^{s}),...,(\textbf{x}_{N_s}^{s},\textbf{y}_{N_s}^{s}) \}$, where $K_s$ is the number of seen categories, $N_s$ is the number of source samples, $\textbf{x}_i^{s} \in \mathcal{R}^d$  is the image feature, $\textbf{y}_i^{s} \in \{ 0,1\}^{K_s}  $ is the one-hot code indicating the class label of $\textbf{x}_i^{s}$ and $d$ is the dimension of image features. In the transductive setting, we are also provided with an unlabeled target dataset $\mathcal{D}^t =\{\textbf{x}_1^{t},...,\textbf{x}_{N_t}^{t} \}$. Note that the target images are from target categories $\mathcal{C}^t=\{c_{1}^{t},...,c_{K_t}^{t} \}$ satisfying $\mathcal{C}^s  \cap \mathcal{C}^t =\o $. Albeit being disjoint, the source and target categories are assumed to share a common semantic space, on which each category $i \in \mathcal{C}^s  \cup \mathcal{C}^t$ is featured by a semantic vector $\textbf{a}_i \in \mathcal{R}^r$, where $r$ is the dimension of semantic vectors. The elements of $\textbf{a}_i \in \mathcal{R}^r$ refer to the semantic attributes of each category. This shared space constitutes the key factor for cross-category transfer. To sum up, our goal is to learn a prediction model which can predict the labels of target images from $\mathcal{C}^t$, given the labeled source data $\mathcal{D}^s$,  the unlabeled target data $\mathcal{D}^t$ and the semantic embeddings $\{\textbf{a}_i\}_{i \in \mathcal{C}^s  \cup \mathcal{C}^t}$ for training.

	\subsection{Connection between ZSL and domain adaptation}
	
	\noindent \textbf{Problem redefinition.} The key idea of TSVN lies in tackling transductive ZSL as a domain adaptation task. To be specific, we adopt the fusion of visual features and semantic attributes $(\textbf{x}_{i}^{s},\textbf{a}_{j}^{s})$, where $i=1,...,N_s$ and $j=1,...,K_s$, to redefine the training samples. Naturally, $\textbf{y}_{ij}^{s}$ is the label of $(\textbf{x}_{i}^{s},\textbf{a}_{j}^{s})$, indicating the similarity/dissimilarity fact of the corresponding semantic-visual fusion. The target domain contains unlabeled semantic-visual fusions $(\textbf{x}_{i}^{t},\textbf{a}_{j}^{t})$, where $i=1,...,N_t$ and $j=1,...,K_t$. By this redefinition, the source and target domains will share the identical task, which leads to a standard unsupervised domain adaptation problem. 
	
	\vspace{0.1cm}
	
	\noindent \textbf{Theoretical insight.} Our proposal also has a strong theoretical interpretation. The previous ZSL approaches usually use the compatibility function $f$ to measure the relationship between  visual features and semantic attributes. In general, $f$ usually  takes the following bilinear form:
	\begin{equation}
		\begin{aligned}
			f(\textbf{x}, \textbf{a};  {\rm W}) =  \textbf{x}^T  {\rm W}  \textbf{a},
		\end{aligned}
	\end{equation}
	where ${\rm W} \in \mathcal{R}^{d \times r }$ is a learnable parameter. It is easy to verify that $f$ can be  equivalently formulated as follows:
	\begin{equation}
		\begin{aligned}
			f(\textbf{x}, \textbf{a};  {\rm u}) = (\textbf{x} \otimes   \textbf{a})^T  {\rm u},
		\end{aligned}
	\end{equation}
	where $ {\rm u} \in \mathcal{R}^{dr}$ is a learnable parameter. Naturally, we can view  $\textbf{x}  \otimes   \textbf{a}$ as a data instance. As a result,  our problem can be treated as predicting the binary labels $y_{ij}$ for the tensor product  of visual features $\textbf{x}_i$ and semantic attributes $\textbf{a}_j$. The binary label $y_{ij} \in \{0,1\}$ indicates whether $\textbf{x}_i$ belongs to the $j$-th category. Hence, the loss function can be formulated as follows:
	\begin{equation}
		\begin{aligned}
			\min_{{\rm u}}  \  \sum_{\textbf{x}_i \in \mathcal{D}^s \cup  \mathcal{D}^t  } \sum_{j \in \mathcal{C}^s \cup  \mathcal{C}^t } \ell(f( \textbf{x}_i,\textbf{a}_{j} ; {\rm u}), y_{ij}) .
		\end{aligned}
	\end{equation}
	With the target domain  unlabeled, transductive ZSL  aims to generalize the prediction ability of $f(\textbf{x}, \textbf{a};  {\rm u}) $ trained with the labeled source domain  to the target  domain. Denote by $\hat{\mathcal{D}}^s=\{\textbf{x}_i  \otimes   \textbf{a}_j | \textbf{x}_i \in \mathcal{D}^s$, $j \in \mathcal{C}^s\}$ and  $\hat{\mathcal{D}}^t=\{\textbf{x}_i  \otimes   \textbf{a}_j | \textbf{x}_i \in \mathcal{D}^t$, $j \in \mathcal{C}^t\}$ the set of semantic-visual fusions from the source and target domains, respectively.  We can see that the distribution discrepancy between $\hat{\mathcal{D}}^s$ and  $\hat{\mathcal{D}}^t$ establishes the bottleneck for knowledge transfer in transductive ZSL, which is also the key point in the standard unsupervised domain adaptation problems. Denote by $\epsilon_{s}(f) $  and $\epsilon_{t}(f) $ the expected error of $f(\cdot;{\rm u})$ on the source and target semantic-visual fusions, respectively.  Denote by $\hat{\mathcal{S}}$ and $\hat{\mathcal{T}}$ the underlying distributions of $\hat{\mathcal{D}}^s$ and $\hat{\mathcal{D}}^t$, respectively.  As in~\cite{theory}, $\epsilon_{t}(f) $ can be bounded as follows: 
	\begin{equation}
		\label{bound}
		\begin{aligned}
			\epsilon_{t}(f) \leq  \epsilon_s(f) + \frac{1}{2} d_{\mathcal{H}}(\hat{\mathcal{S}},\hat{\mathcal{T}}) + \lambda,  \ \forall  f \in  \mathcal{H}
		\end{aligned}
	\end{equation}
	with
	\begin{equation}
		\lambda = \min_{f \in \mathcal{H}} \left [\epsilon_s(f) + \epsilon_t(f) \right ] ,
	\end{equation}
	\begin{equation}
		d_{\mathcal{H}}(\hat{\mathcal{S}},\hat{\mathcal{T}}) = 2 \sup_{f \in \mathcal{H}} \left |  \mathop{\textbf{Pr}}_{\textbf{x} \in \hat{\mathcal{S}}}[f(\textbf{x})=1]   -  \mathop{\textbf{Pr}}_{\textbf{x} \in \hat{\mathcal{T}}}[f(\textbf{x})=1]  \right | ,
	\end{equation}
	where $\lambda$ is the ideal expected error, $d_{\mathcal{H}}(\hat{\mathcal{S}},\hat{\mathcal{T}})$ is the $\mathcal{H}$-divergence between $\hat{\mathcal{S}}$ and $\hat{\mathcal{T}}$. Hence, it is clear that  reducing the distribution discrepancy of semantic-visual fusions can improve the performance of transductive ZSL.
	
	\begin{figure*}[t]
		\centering
		{\includegraphics[width=0.98\textwidth,trim=0 0 100 0,clip]{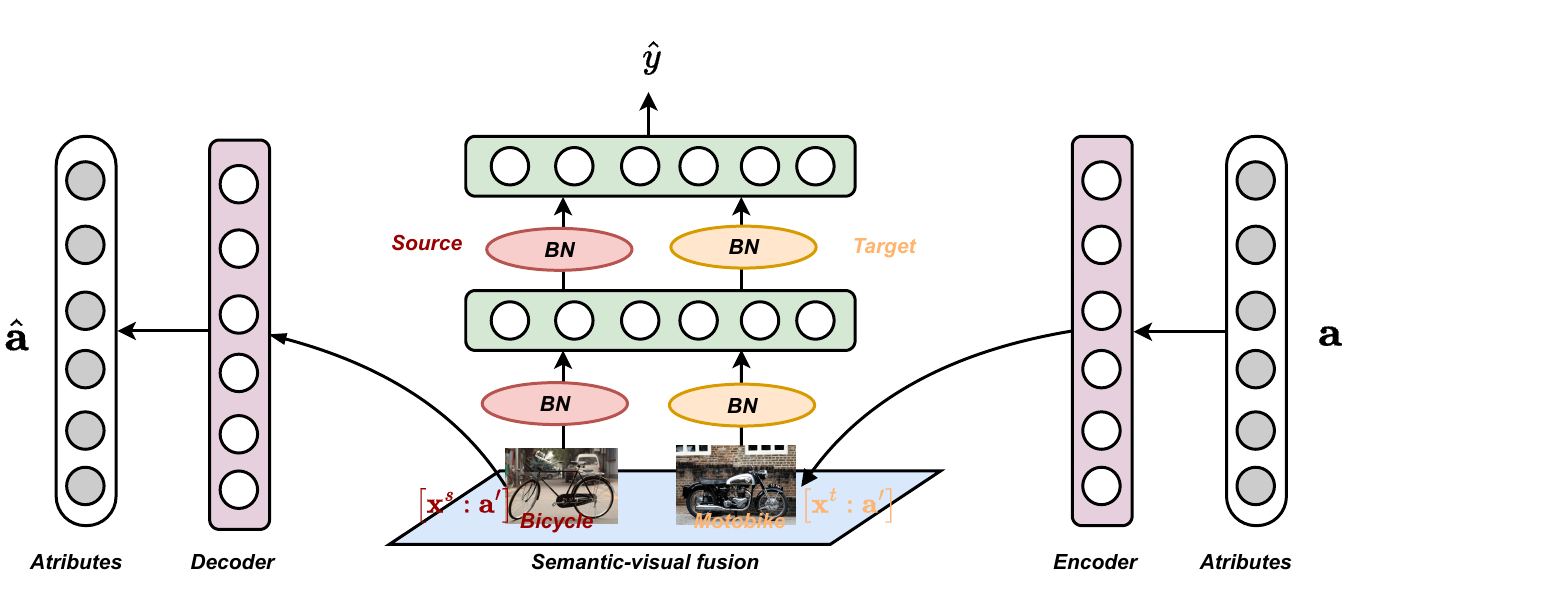}}
		\caption{The overall architecture of our proposed TSVN. The encoder projects the semantic attribute vector $\textbf{a}$ into a transformed representation $\textbf{a}'$. The multi-layer neural network takes the concatenated semantic-visual fusions $\left [ \textbf{x}:\textbf{a}' \right ]$ as inputs and outputs their similarity score $\hat{\textbf{y}}$. The decoder takes the transformed semantic representation $\textbf{a}'$ as input and outputs the reconstructed semantic attribute vector $\hat{\textbf{a}}$. In each batch normalization layer, two BN units are used to normalize the mini-batches from the source and target domains separately.} 
		\label{model}
	\end{figure*}

	\subsection{Model overview}
	
	We use a hierarchical metric network as the base model for the transformed domain adaptation problem. Since the dimension of the tensor product $\textbf{x} \otimes \textbf{a}$ can be very high, we concatenate the visual features and semantic attributes directly to construct the semantic-visual fusions. Moreover, to increase the flexibility, we use an encoder to project the semantic attribute $\textbf{a}$ into a transformed representation. Denote by $\textbf{a}' \in \mathcal{R}^{r'}$ and $\left [ \textbf{x}:\textbf{a}' \right ] \in \mathcal{R}^{d+r'}$ the transformed semantic attribute representation and the concatenated semantic-visual fusion, respectively.  We then use a hierarchical metric network shared by both two domains to predict the relation score of each concatenated semantic-visual fusion. The overall architecture of our model is clearly displayed in Fig. \ref{model}. By using a hierarchical metric network instead of directly measuring their distance over the embedding space, we can more accurately capture the relation score of each semantic-visual fusion. Following \cite{kodirov2017semantic}, we incorporate a decoder for reconstructing the original semantic attributes, in order to enforce the transformed semantic representations $\textbf{a}'$  to preserve their original semantic information. The reconstruction module is implemented by the L2 norm:
	\begin{equation}
		\begin{aligned}
			d_i =    \left \|   \textbf{a}_i -  \hat{\textbf{a}}_i \right \|^{2} ,
		\end{aligned}
	\end{equation}
	where $\hat{\textbf{a}}_i $ is the output of the decoder. To align the distributions of semantic-visual fusions from the source and target domains, we incorporate two BN modules at each layer to maintain different normalization statistics for two domains during the training phase. Additionally, we incorporate an entropy minimization term as  reasonable  regularization for the  unlabeled semantic-visual fusions from the target domain.
	
	There may exist a potential problem, {i.e.}, how to build mini-batches for training since the semantic-visual fusions are extremely imbalanced.  For the source domain, we randomly select a fixed-size batch of images from $\mathcal{D}^s$, and then use categories existing in the current mini-batch to construct semantic-visual fusions for training. For the target domain, we will also select a  mini-batch of images from $\mathcal{D}^s$, but all the categories from $\mathcal{C}^t$ will be used to construct semantic-visual fusions since the target images do not have  labels.
	
	\subsection{Distribution shift alignment}
	This work reduces the distribution discrepancy based on the recent advances of batch normalization. In this part, we first introduce the background of batch normalization. Then we present how to utilize batch normalization for distribution alignment. 
	
	\vspace{0.1cm}
	
	\noindent \textbf{Preliminary of batch normalization.}  Batch normalization aims to keep the distribution of  mini-batches unchanged across  the intermediate layers of deep models. In this way, the training algorithm of deep neural networks can obtain more stable gradients to update the network parameters.  To be specific,  BN normalizes the mini-batches with {zero} mean and {one} variance at each layer of deep neural networks.  In each BN layer, we firstly calculate the mean value {$\mu$} and variance value {$\sigma^{2}$} of mini batches:
	\begin{equation}
		\begin{aligned}
			\mu^{k}= & \frac{1}{m} \sum_{i=1}^{m} x_{i}^{k}, \\
			\left(\sigma^{2}\right)^{k}= & \frac{1}{m} \sum_{i=1}^{m}\left(x_{i}^{k}-\mu^{k}\right)^{2}, 
		\end{aligned}
	\end{equation}
	where $x_{i}^{k}$ is the \emph{k}-th element of the intermediate activation of the \emph{i}-th sample in the current mini-batch, $m$ is the batch size. Then we normalize the intermediate activations to have {zero} mean and {one} variance:
	\begin{equation}
		\begin{aligned}
			\hat{x}^{k}=\frac{x^{k}-\mu^{k}}{\sqrt{\left(\sigma^{2}\right)^{k}+\epsilon}},
		\end{aligned}
	\end{equation}
	where {$\hat{x}^{k}$ is the normalized intermediate activation} and $\epsilon$ is the parameter introduced to avoid the problem of underflow. Moreover, in order to guarantee the flexility of each intermediate layer,  learnable scale and shift parameters, $\gamma^{k}$ and $\beta^{k}$,  are further used to transform the normalized activation $\hat{x}^{k}$:
	\begin{equation}
		\begin{aligned}
			z^{k}=\gamma^{k} \hat{x}^{k}+\beta^{k}.
		\end{aligned}
	\end{equation}
	Note that each BN layer will retain a set of global normalization statistics $\{\mu^{k}_g, \sigma^{k}_g \}$ using the moving average mechanism:
	\begin{equation}
		\begin{aligned}
			\mu^{k}_g = & \alpha \mu^{k}_g+ (1-\alpha) \mu^{k}, \\
			\sigma^{k}_g= & \alpha \sigma^{k}_g + (1-\alpha) \sigma^{k}, 
		\end{aligned}
	\end{equation} 
	where $\mu^{k}$ and $ \sigma^{k}$ are the sample mean and sample standard deviation of the current mini-batch. In the testing phase, the BN layers use the global normalization statistics to normalize the testing data.
	
	\vspace{0.1cm}
	
	\noindent \textbf{Domain-specific batch normalization.}  Although the standard batch normalization enables the deep model to obtain similar data distribution across different layers, it cannot lead to similar data distribution across different domains.  In order to make the source and target data have similar distribution, we propose to incorporate two batch normalization units at each BN layer. This design enables deep models to separately normalize the mini-batches of different domains with {zero} mean and {one} variance.
	
	Specifically, each  BN layer will firstly calculate the mean and variance values for mini-batches of the source and target domains separately: 
	\begin{equation}
		\begin{aligned}
			\left(\sigma^{2}\right)^{s(k)}= & \frac{1}{m} \sum_{i=1}^{m}\left(x_{i}^{s(k)}-\mu^{s(k)}\right)^{2}, \\
			\left(\sigma^{2}\right)^{t(k)}= & \frac{1}{m} \sum_{i=1}^{m}\left(x_{i}^{t(k)}-\mu^{t(k)}\right)^{2}, \\
			\mu^{s(k)}=  \frac{1}{m} &  \sum_{i=1}^{m} x_{i}^{s(k)},  \ \mu^{t(k)}=  \frac{1}{m} \sum_{i=1}^{m} x_{i}^{t(k)}. \\
		\end{aligned}
	\end{equation}
	As a result, the source and target data can be normalized by  their domain-specific normalization statistics:
	\begin{equation}
		\begin{aligned}
			\hat{x}^{s(k)}=\frac{x^{s(k)}-\mu^{s(k)}}{\sqrt{\left(\sigma^{2}\right)^{s(k)}+\epsilon}}, \\
			\hat{x}^{t(k)}=\frac{x^{t(k)}-\mu^{t(k)}}{\sqrt{\left(\sigma^{2}\right)^{t(k)}+\epsilon}}.
		\end{aligned}
	\end{equation}
	After that,  the source and target data are transformed by the identical scale and shift parameters: 
	\begin{equation}
		\begin{aligned}
			z^{s(k)}= & \gamma^{k} \hat{x}^{s(k)}+\beta^{k}, \\
			z^{t(k)}= & \gamma^{k} \hat{x}^{t(k)} +\beta^{k}.
		\end{aligned}
	\end{equation}
	The  detailed calculating process within the domain-specific BN layer is shown in Algorithm \ref{ABN}. Note that each BN layer will also retain two sets of global normalization statistics, {i.e.}, $\{\mu^{s(k)}_g, \sigma^{s(k)}_g  \}$ and $\{ \mu^{t(k)}_g, \sigma^{t(k)}_g  \}$, to normalize the source and target testing data separately in the testing stage.
	
	\begin{algorithm}[t]
		\caption{Domain-specific  batch normalization}\label{ABN}
		\LinesNumbered  
		\KwIn{ Dimension of  intermediate layer $K$; batch size $m$; mini-batch of  intermediate vectors $\{ \textbf{x}^s_i \}_{i=1}^m$ and $\{ \textbf{x}^t_i \}_{i=1}^m$,  where $\textbf{x}^{s}_i, \textbf{x}^{t}_i \in \mathcal{R}^{ K}$; scale and shift parameters: $\gamma, \beta \in \mathcal{R}^{K}$.} 
		
		\KwOut{ normalized intermediate vectors   $\{ \textbf{z}^s_i \}_{i=1}^m$ and $\{ \textbf{z}^t_i \}_{i=1}^m$, where $\textbf{x}^{s}_i, \textbf{x}^{t}_i \in \mathcal{R}^{ K}$. }
		
		Calculate the mean values of each element: $\mu^{t(k)}=\frac{1}{m} \sum_{i=1}^{m} x_{i}^{t(k)}, \mu^{s(k)}=\frac{1}{m} \sum_{i=1}^{m} x_{i}^{s(k)}$.     \\
		
		Calculate the variance values of each element: $\left(\sigma^{2}\right)^{s(k)}=\frac{1}{m} \sum_{i=1}^{m}\left(x_{i}^{s(k)}-\mu^{s(k)}\right)^{2}, \left(\sigma^{2}\right)^{t(k)}=\frac{1}{m} \sum_{i=1}^{m}\left(x_{i}^{t(k)}-\mu^{t(k)}\right)^{2}$. \\
		
		Normalize $\textbf{x}^{s}$ and $\textbf{x}^{t}$ separately: $\hat{x}^{s(k)}=\frac{x^{s(k)}-\mu^{s(k)}}{\sqrt{\left(\sigma^{2}\right)^{s(k)}+\epsilon}}, \hat{x}^{t(k)}=\frac{x^{t(k)}-\mu^{t(k)}}{\sqrt{\left(\sigma^{2}\right)^{t(k)}+\epsilon}}$.   \\
		
		Transform the normalized vector by the scale and shift parameters: $z^{s(k)}=\gamma^{k} \widehat{x}^{s(k)}+\beta^{k}, z^{t(k)}=\gamma^{k} \widehat{x}^{t(k)}+\beta^{k}$ . \\ 
	\end{algorithm}
	
	The domain-specific batch normalization approach suits our redefined domain adaptation problem well. Although the number of positive fusions is very small in each mini-batch, the moving average mechanism of batch normalization enables to retain their effect in feature alignment. Compared to MMD or domain adversarial training which aligns distributions over mini-batches, the problem of data imbalance can be alleviated. Moreover, our design is very easy-to-implement, without introducing extra hyper-parameters or mini-max problems which are often difficult to be optimized. Compared with \cite{li2018adaptive}, the domain-specific batch normalization design allows us to train the network by  the target data without introducing bias in the normalization statistics.  
	
	\vspace{0.1cm}
	
	\noindent \textbf{Theoretical insight.} From Eq.~\ref{bound}, it is clear that the distribution discrepancy between $\hat{\mathcal{S}}$ and  $\hat{\mathcal{T}}$ builds the main bottleneck for knowledge transfer in domain adaptation. In Theorem 1, we show the theoretical interpretation of domain-specific batch normalization.
	
	\begin{theorem}
		Domain-specific batch normalization is trying to  reduce the upper bound $d_{\mathcal{H}}(\hat{\mathcal{S}},\hat{\mathcal{T}})$.
	\end{theorem}
	
	\begin{proof}
		It is clear that $d_{\mathcal{H}}(\hat{\mathcal{S}},\hat{\mathcal{T}})$  satisfies the triangular inequation:
		\begin{equation}
			\label{triangular}
			\begin{aligned}
				d_{\mathcal{H}}(\mathcal{S},\mathcal{T}) =& 2 \sup_{f \in \mathcal{H}} \left |  \mathop{\textbf{Pr}}_{\textbf{x} \in \hat{\mathcal{S}}}[f(\textbf{x})=1]   -  \mathop{\textbf{Pr}}_{\textbf{x} \in \hat{\mathcal{T}}}[f(\textbf{x})=1]  \right | \\
				\leq & 2 \sup_{f \in \mathcal{H}} \left |  \mathop{\textbf{Pr}}_{\textbf{x} \in \hat{\mathcal{S}}}[f(\textbf{x})=1]   - \mathop{\textbf{Pr}}_{\textbf{x} \in {\mathcal{P}}}(\textbf{x})  \right | + \\
				&     2 \sup_{f \in \mathcal{H}} \left |  \mathop{\textbf{Pr}}_{\textbf{x} \in \hat{\mathcal{T}}}[f(\textbf{x})=1]   -  \mathop{\textbf{Pr}}_{\textbf{x} \in {\mathcal{P}}}(\textbf{x}) \right | \\
				= & d_{\mathcal{H}}(\hat{\mathcal{S}},\mathcal{P}) + d_{\mathcal{H}}(\hat{\mathcal{T}},\mathcal{P}).
			\end{aligned}
		\end{equation}
		The distribution $\mathcal{P}$ satisfies the condition that $\mathcal{P}_k \sim \mathcal{N}(0,1)$, where  $\mathcal{P}_k$ is the marginal distribution of $\mathcal{P}$ at each dimension. Denote by $\hat{\mathcal{S}}_k$ and $\hat{\mathcal{T}}_k$ the marginal distributions of each dimension of the source and target domains, respectively. In each domain, the domain-specific batch normalization unit enforces the marginal distribution of each dimension of the data instance to  approximate the normal distribution: $\hat{\mathcal{S}}_k \rightarrow \mathcal{N}(0,1)$ and $ \hat{\mathcal{T}}_k  \rightarrow \mathcal{N}(0,1)$. As a result, the last two terms of Eq. \ref{triangular}  will be enforced to be close to  0. Hence, domain-specific batch normalization gives a tighter upper bound for the  expected error $\epsilon_t(f)$.
	\end{proof}
	
	\subsection{Practical entropy regularization}
	Furthermore, we also incorporate practical regularization for the prediction of  unlabeled semantic-visual fusions from the target domain. We regularize the metric network by leveraging the property that a target image can only relate to one category from $\mathcal{C}^t$. Through enforcing the metric network to reach this property, we can strengthen the network's generalization ability for the target domain. To implement this, we design an entropy minimization loss for the target semantic-visual fusions. To be specific, for a target image $\textbf{x}_{i}^{t}$, we collect the metric network's outputs of each semantic-visual fusion as a vector: $\textbf{Z}_{i}^{t} = \left [\textbf{z}_{i1}^{t}, ..., \textbf{z}_{iK_t}^{t}  \right ]$, where $\textbf{z}_{ij}^{t}$ is the output for the fusion of $\textbf{x}_{i}^{t}$ and $\textbf{a}_{j}^{t}$. Note that $\textbf{z}_{ij}^{t}$ is collected before the sigmoid layer for avoiding the problem of gradient vanishing. Then we use a softmax operation to process $\textbf{Z}_{i}^{t}$, resulting in a probability vector $\textbf{P}_{i}^{t} = \left [\textbf{p}_{i1}^{t}, ..., \textbf{p}_{iK_t}^{t}  \right ]$ that describes which category the target image $\textbf{x}_{i}^{t}$ belongs to.  To reach the property as indicated above, we propose to minimize the entropy of that probability vector since small entropy implies low-density separation between categories \cite{krause2010discriminative}. Formally, the entropy loss is defined as:
	\begin{equation}
		\begin{aligned}
			{\rm H}(\textbf{P}_{i}^{t})   
			=  -  \sum \limits_{j=1}^{K_t} \textbf{p}_{ij}^{t} \log \textbf{p}_{ij}^{t} .
		\end{aligned}
	\end{equation}
	
	\subsection{Training}
	From our definition of the redrawn domain adaptation problem, only the source semantic-visual fusions have similarity/dissimilarity  labels. Hence, using $\hat{\textbf{y}}_{ij}^{s}$ to denote the metric network's prediction for the fusion of $\textbf{x}_{i}^{s}$ and $\textbf{a}_{j}^{s}$, we can train the metric network via a mean squared error loss in the source domain:
	\begin{footnotesize}
		\begin{equation}
			\begin{aligned}
				\mathcal{L}_{\rm pre}
				= & - \frac{1}{N_s}  \frac{1}{K_s} \sum \limits_{i=1}^{N_s} \sum \limits_{j=1}^{K_s}  (\textbf{y}_{ij}^{s}-\hat{\textbf{y}}_{ij}^{s})^{2}.
			\end{aligned}
		\end{equation}
	\end{footnotesize}
	The entropy regularization loss defined for target semantic-visual fusions is formalized as:
	\begin{equation}
		\begin{aligned}
			\mathcal{L}_{\rm ent} =   \frac{1}{N_t} \sum \limits_{i=1}^{N_t} {\rm H}(\textbf{P}_{i}^{t})   .   \\
		\end{aligned}
	\end{equation}
	The reconstruction loss is computed for the semantic attributes of both source and target categories:
	\begin{equation}
		\begin{aligned}
			\mathcal{L}_{\rm rec}
			=  \frac{1}{K_s} \sum \limits_{i=1}^{K_s} d_i^s  + \frac{1}{K_t} \sum \limits_{i=1}^{K_t} d_i^t .
		\end{aligned}
	\end{equation}
	
	In conclusion, with the above sub-objectives, our final objective function is formulated as follows:
	\begin{equation}
		\min  \mathcal{L}_{\rm pre} + \lambda_{\rm ent} \mathcal{L}_{\rm ent} + \lambda_{\rm rec} \mathcal{L}_{\rm rec},
	\end{equation}
	where $\lambda_{\rm rec}$ and $\lambda_{\rm ent}$ are hyper-parameters that weigh the importance of the corresponding terms. It is worth noting that our design for distribution shift alignment does not introduce any loss functions, as well as extra hyper-parameters.  In optimization, the encoder and decoder for semantic attributes  receive gradients from all the three loss terms, while the metric network receives gradients from the first two loss terms.
	
	\subsection{Prediction}
	After the training phase is finished, we can obtain the metric network's prediction $\hat{\textbf{y}}_{ij}^{t}$ for target domain semantic-visual fusions. Specifically, $\hat{\textbf{y}}_{ij}^{t}$ reveals the relation score of the fusion of image $\textbf{x}_{i}^{t}$ and attribute $\textbf{a}_{j}^{t}$. Hence, we can predict the label of target domain image $\textbf{x}_{i}^{t}$ as the class with the largest relation score: 
	\begin{equation}
		\ k^* = \mathop{\rm argmax}_{j \in \mathcal{C}^{t}} \ \hat{\textbf{y}}_{ij}^{t} .
	\end{equation}

	\section{Experiments}
	\label{expirements}
	\subsection{Datasets}
	We conduct experiments on four standard ZSL datasets: 1) attribute-Pascal-Yahoo (aPY) \cite{farhadi2009describing}; 2) Animal with Attribute (AwA) \cite{xian2018zero}; 3)  Caltech-UCSD Birds 200-2011 (CUB) \cite{wah2011caltech}; 4) SUN-Attribute (SUN) \cite{patterson2012sun}. The
	aPY dataset has 12,695 images of 20 classes  from Pascal  and 2,644 images of 12 classes from Yahoo. The images of aPY are labeled by 64 dim binary vectors for denoting the semantic attributes. For aPY, the source domain includes 20 categories and the target domain has 12 ones. AwA contains 37,322 images from 50 categories. For AwA, the source and target domains have 40 and 10 categories, respectively. The images in AwA are labeled by 85-dimensional semantic attributes. The CUB dataset contains 11,788 images over 200 bird species, in which 150 species are used as the source categories and the rest 50 as the target. The images in CUB are labeled by  312 dimensional continuous vectors. The SUN dataset contains 14,140 images over 717 categories, with each image labeled by a 102 dimensional attribute vector. For SUN, the source and target domains contain 645 and 72 categories, respectively.  
	
	\subsection{Implementation details}
	Our method is implemented by the PyTorch framework. In all the experiments, the image embeddings are 2048 dimensional features obtained from the top-layer units of ResNet-101. The ResNet-101 is pre-trained on ImageNet with 1K classes. Both the encoder and decoder networks include one hidden layer with 1250 units.  The metric network contains two fully connected layers with 1250  hidden units.  In our experiments, Adam is utilized as the optimizer. We set the initial learning rate to $10^{-5}$.  The maximum iteration number is 50,000 and the batch size is set to 32. The trade-off parameters $\lambda_{\rm rec}$ and $\lambda_{\rm ent}$ are set to $10^{-5}$ and $10^{-9}$, respectively.  Our experimental results are insensitive to their values. Like the existing {state-of-the-art} transductive ZSL methods \cite{kodirov2015unsupervised,xian2018zero}, we adopt the label propagation step to refine the final prediction results.
	
	In our experiments, we employ both the standard splits (SS) and the proposed splits  (PS) from \cite{xian2018zero} for fair comparisons. We use the Mean Class Accuracy (MCA) as our evaluation metric, which is defined as follows:
	\begin{equation}
		{\rm MCA} = \frac{1}{|\mathcal{C}^t|} \sum \limits_{y \in \mathcal{C}^t} {acc}_{y},
	\end{equation}
	where ${acc}_{y}$ denotes the prediction accuracy over the test data of class $y$. The experimental results  are averaged over five independent trials. Note that we do not conduct experiments for the generalized  setting since it is ill-posed for transductive ZSL: 1) The transductive setting means that the unlabeled data of the training phase will also be the data for testing, which suits for ZSL as well \cite{xian2018zero}. Actually, this is problematic for the generalized ZSL setting, since the model has already known that the unlabeled data are from the target categories during training. 2) Since both the source and target images are available for training, we can naively train a binary classifier for predicting the domain label of testing data. Hence, the performance will mainly rely on the capacity of the binary classifier.
	
	\subsection{Baselines}
	
	\begin{table}[t]
		\renewcommand\arraystretch{1.3}
		\fontsize{9.0}{10.2} \selectfont
		\caption{Comparisons with the existing state-of-the-art ZSL baselines in each setting. The notation $\mathscr{I}$ denotes the inductive baselines and $\mathscr{T}$ denotes the transductive baselines.}
		\vspace{0.2cm}
		\centering
		\resizebox{\columnwidth}{!}{
			\begin{tabular}{c|c|rr|rr|rr|rr}
				\hline
				\multicolumn{1}{l|}{}     & \multicolumn{1}{l|}{} & \multicolumn{2}{c|}{CUB}                         & \multicolumn{2}{c|}{aPY}                         & \multicolumn{2}{c|}{AwA}                        & \multicolumn{2}{c}{SUN}                         \\
				\multicolumn{1}{l|}{}     & Model                & \multicolumn{1}{c}{PS} & \multicolumn{1}{c|}{SS} & \multicolumn{1}{c}{PS} & \multicolumn{1}{c|}{SS} & \multicolumn{1}{c}{PS} & \multicolumn{1}{c|}{SS} & \multicolumn{1}{c}{PS} & \multicolumn{1}{c}{SS} \\ \hline
				\multirow{14}{*}{$\mathscr{I}$}       & ALE                  & 54.9\%                 & 53.2\%                 & 39.7\%                 & 30.9\%                 & 62.5\%                 & 80.3\%                 & 58.1\%                 & 59.1\%                 \\
				& SAE                  & 33.3\%                 & 33.4\%                 & 8.3\%                  & 8.3\%                  & 54.1\%                 & 80.7\%                 & 40.3\%                 & 42.4\%                 \\
				& DCN                  & 56.2\%                 & 55.6\%                 & 43.6\%                 & -                      & 65.2\%                 & 82.3\%                 & 61.8\%                 & \textbf{67.4}\%                 \\
				& SRN                  & 56.0\%                 & -                      & 38.4\%                 & -                      & -                      & -                      & 61.4\%                 & -                      \\
				& SP-AEN               & 55.4\%                 & -                      & 24.1\%                 & -                      & 58.5\%                 & -                      & 59.2\%                 & -                      \\
				& SE-ZSL               & 59.6\%                 & 60.3\%                 & -                      & -                      & 69.5\%                 & 83.8\%                 & 63.4\%                 & 64.5\%                 \\
				& f-CLSWGAN                  & 57.3\%                 & -                      & -                      & -                      & 68.2\%                 & -                      & 60.8\%                 & -                      \\
				& LSE                  & -                 & 55.4\%                      & -                      & \textbf{47.6\%}                      & 68.2\%                 & -                      & 60.8\%                 & -                      \\ 
				& LAF & 58.1\% & - & 47.1\% & - & 67.2\% & - & \textbf{65.3\%} & - \\ 
				& SZSL & 60.1\% & - & 45.6\% & - & 72.5\% & - & 64.1\% & - \\
				& SRSA & 59.9\% & - & 43.3\% & - & 68.3\% & - & 64.3\% & - \\
				& EXEM & 58.0\% & - & - & - & 64.6\% & - & 62.9\% & - \\
				& VSOP & 52.6\% & - & 45.3\% & - & 74.1\% & - & 60.8\% & - \\
				& LSG & 52.9\% & - & 30.4\% & - & 61.1\% & - & 53.4\% & - \\
				& DRFCG & 60.6\% & - & 42.8\% & - & 72.4\% & - & 63.9\% & - \\\hline
				\multirow{8}{*}{$\mathscr{T}$}   & ALE-T	& 54.5\%	& 59.4\%	& 46.4\%	& 40.7\%	& 70.6\%	& 60.6\%	& 55.4\%	& 42.2\% \\
				& UDA                  & -                      & 39.5\%                 & -                      & -                      & -                      & -                  & -                      & -                      \\
				& SMSL                 & -                      & -                      & -                      & 39.0\%                 & -                      & -                      & -                      & -                      \\
				& TMVL                 & -                      & 47.9\%                 & -                      & -                      & -                      & -                      & -                      & -                      \\
				& LisGAN               & 58.8\%                 & -                      & 43.1\%                 & -                      & 70.6\%                 & -                      & 61.7\%                 & -                      \\
				& TSTD               & -                 & 58.2\%                      & -                 & -                      & -                 & -                      & -                 & -                      \\
				& ZeroNAS               & \textbf{62.6\%}                 & -                     & -                 & -                      & 71.3\%                 & -                      & 61.4\%                 & -                      \\
				\hline
				\multicolumn{1}{l|}{ours} & TSVN                 & 61.0\%        & \textbf{60.9\%}        & \textbf{53.3\%}        & 46.3\%                 & \textbf{74.6\%}        & \textbf{93.2\%}        & 64.7\%        & 66.7\%                 \\\hline
		\end{tabular}}
		\label{compare}
	\end{table}
	
	To demonstrate the effectiveness of TSVN, we compare TSVN with a number of baseline methods under two different settings: i)  inductive Zero-Shot Learning baselines  (i.e., ALE~\cite{akata2013label}, SAE~\cite{kodirov2017semantic}, DCN~\cite{liu2018generalized}, SRN~\cite{annadani2018preserving}, SP-AEN~\cite{chen2018zero}, SE-ZSL~\cite{kumar2018generalized}, f-CLSWGAN~\cite{xian2018feature}, LSE~\cite{LSE}, LAF~\cite{liu2020label}, SZSL~\cite{shen2021spherical}, SRSA~\cite{liu2022zero}, EXEM~\cite{changpinyo2020classifier}, VSOP~\cite{wu2021joint}, LSG~\cite{xu2021semi}, and DRFCG~\cite{liu2022learning}); ii) transductive Zero-Shot Learning baselines (i.e., ALE-T~\cite{akata2015label}, UDA~\cite{kodirov2015unsupervised}, SMSL~\cite{guo2016transductive}, TMVL~\cite{fu2015transductive}, LisGAN~\cite{li2019leveraging}, TSTD~\cite{TSTD}, and ZeroNAS~\cite{yan2021zeronas}).

	\begin{table}[]
		\renewcommand\arraystretch{1.3}
		\fontsize{9.0}{10.2} \selectfont
		\caption{Ablation study for the contribution of each design over the PS setting.  ``Baseline'' indicates the metric network trained with only the semantic-visual fusions from the source domain; ``\emph{-2BN}'' indicates using only one batch normalization unit in each BN layer. ``\emph{-reconstruct}'' and ``\emph{-entropy}'' indicate removing $\mathcal{L}_{\rm rec}$ and $\mathcal{L}_{\rm ent}$ from the overall objective, respectively; the last four lines indicate replacing the distribution alignment approach with DANN~\cite{tzeng2017adversarial}, MMD~\cite{long2015learning}, CMD~\cite{zellinger2019robust}, and EMD\cite{shen2018wasserstein}, respectively.}
		\vspace{0.2cm}
		\centering
		\setlength{\tabcolsep}{3.5mm}{
			\begin{tabular}{l|r|r|r|r}
				\hline
				& \multicolumn{1}{c|}{CUB} & \multicolumn{1}{c|}{aPY} & \multicolumn{1}{c|}{AwA} & \multicolumn{1}{c}{SUN} 
				\\ \hline
				Baseline      & 58.0\%                  & 35.9\%                  & 62.4\%                   & 63.1\%                  \\
				\hline
				Full model    & 61.0\%                  & \textbf{53.3\%}                  & \textbf{74.6\%}                   & \textbf{64.7\%}                  \\
				\ \  \emph{-2BN}         & 59.4\%                  & 49.4\%                  & 67.7\%                   & 64.1\%                  \\
				
				\ \   \emph{-reconstruct} & 59.5\%                  & 46.7\%                  & 65.0\%                   & 62.6\%                  \\
				
				\ \  \emph{-entropy}     & \textbf{61.8\%}                  & 49.1\%                  & 68.4\%                   & 64.1\%                  \\ 
				\hline
				DANN        & 57.4\%                  & 49.8\%                  & 68.5\%                   & 63.5\%                  \\
				MMD           & 59.4\%                  & 49.4\%                  & 69.0\%                   & 62.9\%                  \\
				CMD           & 57.4\%                  & 38.6\%                  & 64.2\%                   & 62.4\%                  \\
				EMD           & 60.4\%                  & 45.2\%                  & 64.5\%                   & 64.1\%                  \\
				\hline
		\end{tabular}}
		\label{ablation}
	\end{table}
	
	\subsection{Performance comparison}
	The experimental results of our method and the existing {state-of-the-art} ZSL methods are reported in Table \ref{compare}. For more convincing comparisons, we report the results of both the SS and PS settings.  It is clear that: 1) in general, the transductive baselines usually have better performance than the inductive baselines by introducing unlabeled data from the target categories for training; 2) our performance is on par with or better than the existing {state-of-the-art} baselines. Specifically, with label propagation as the only transductive design, ALE-T cannot sufficiently exploit the potential semantic relations between the source and target categories. Compared with UDA or SMSL, our model can more closely associate the source and target categories  by aligning their distributions over multiple layers rather than merely utilizing the direct relation of semantic representation. For TMVL, although it exploits multiple intermediate semantic representations, its performance is still unsatisfactory. It is also worthy noting that our method is very easy-to-implement, without introducing tedious EM optimization procedures like SMSL. Compared with LisGAN, our method does not need to solve the mini-max problems which are often difficult to optimize. 
	
	\subsection{Analysis}
	
	\begin{figure*}[t]
		\centering
		\subfloat[]{\includegraphics[width=0.24\textwidth,trim=19 21 63 70,clip]{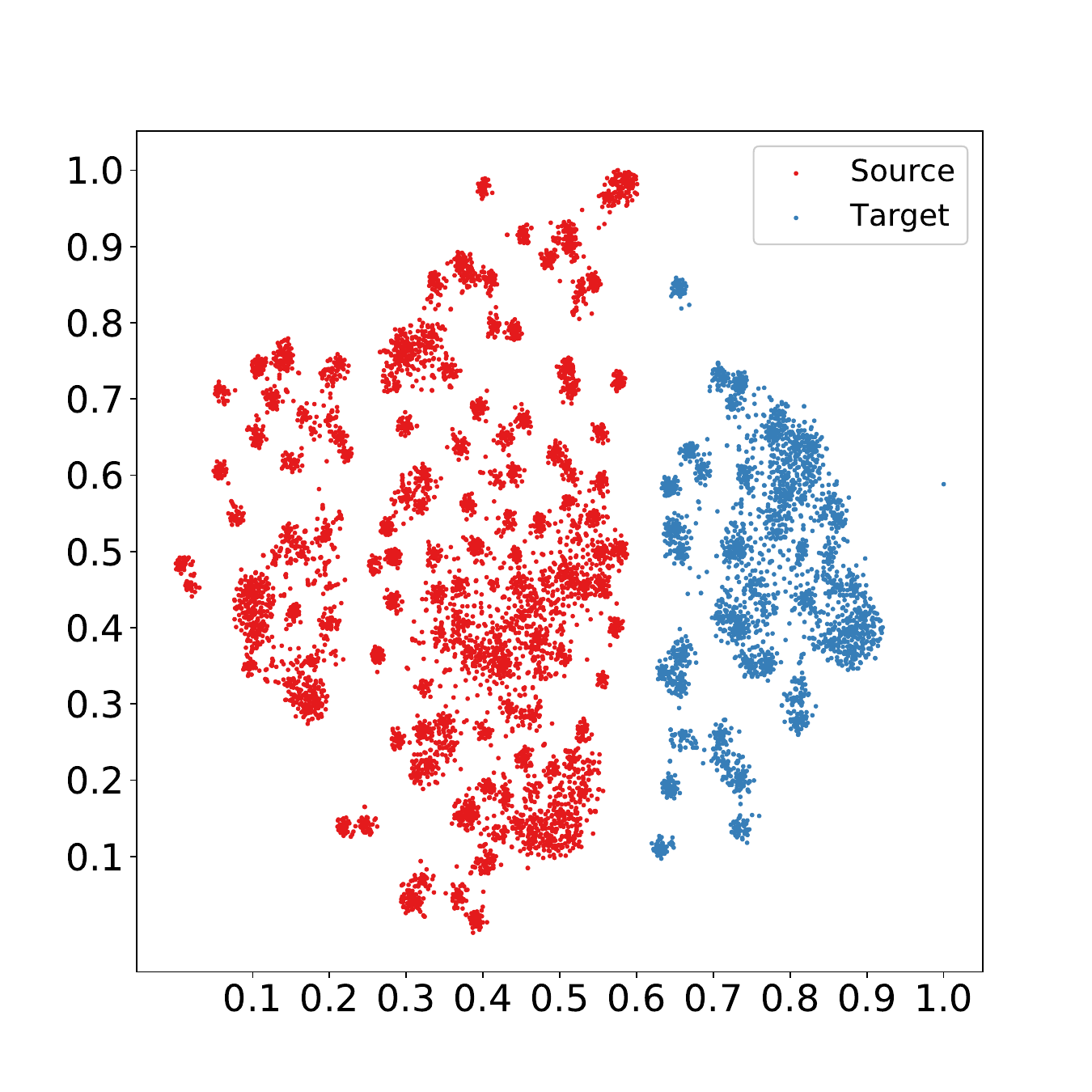}}
		\hfil
		\subfloat[]{\includegraphics[width=0.24\textwidth,trim=19 21 63 70,clip]{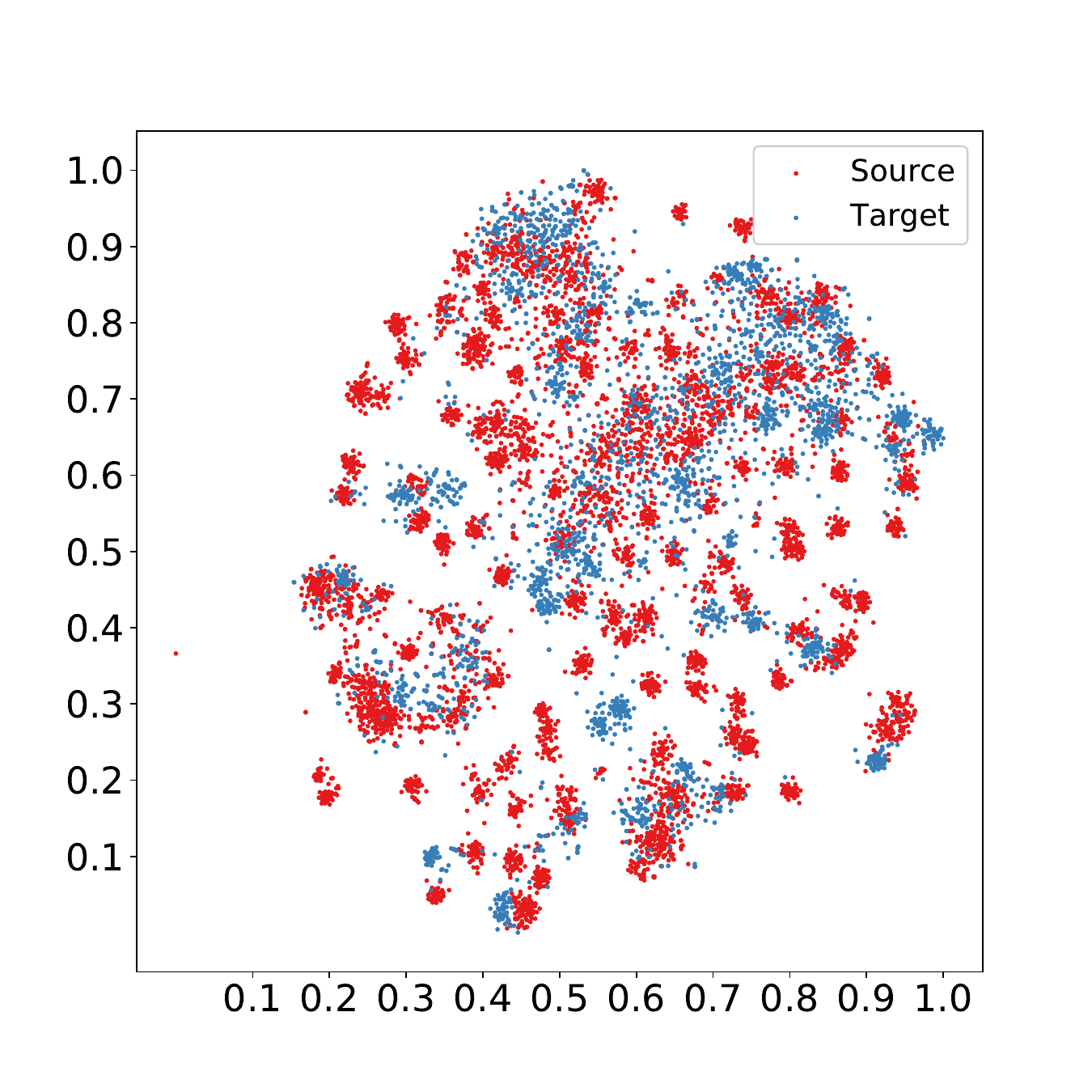}}
		\hfil
		\subfloat[]{\includegraphics[width=0.24\textwidth,trim=19 21 63 70,clip]{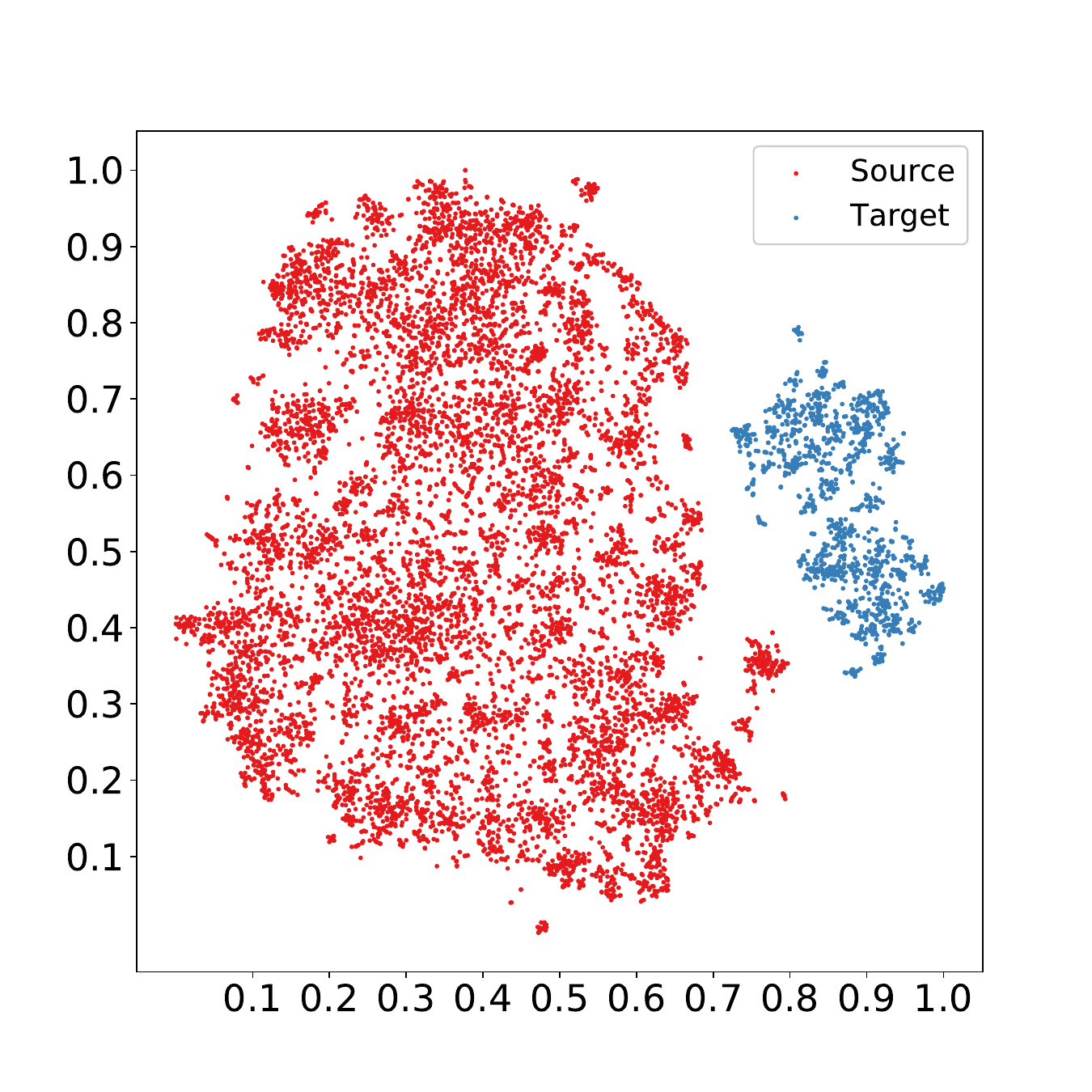}}
		\hfil
		\subfloat[]{\includegraphics[width=0.24\textwidth,trim=19 21 63 70,clip]{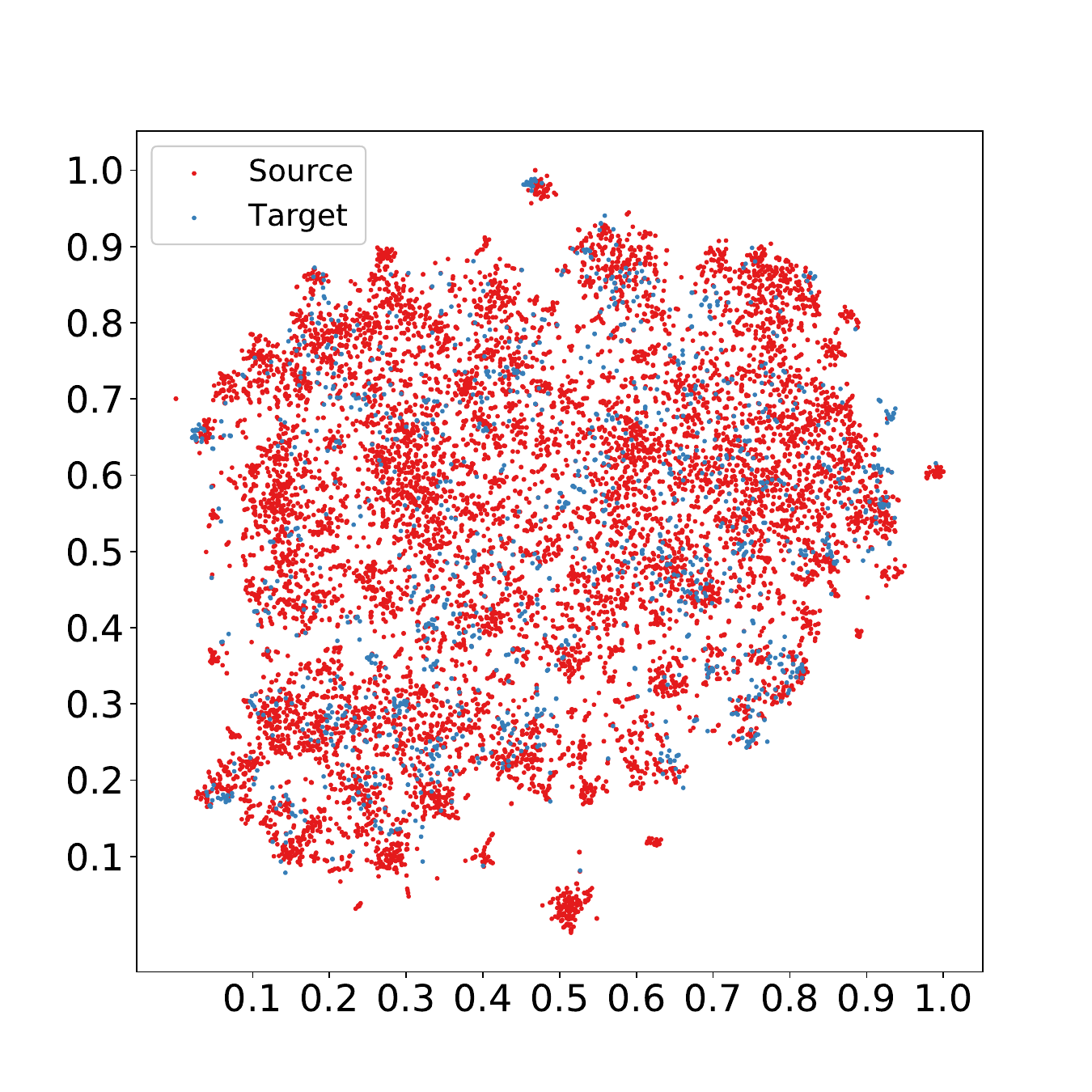}}
		\caption{The t-SNE visualization for intermediate layers of the metric network, in which (a) NoAdapt for CUB, (b) 2BN for CUB, (c) NoAdapt for SUN, and (d) 2BN for SUN.}
		\label{visual}
	\end{figure*}

	As displayed in Table~\ref{ablation}, we further conduct the ablation study to demonstrate the contributions of each design in TSVN. The first row represents the metric network trained with only the semantic-visual fusions from source domain. From the next four rows, it is clear that all the key points designed in our approach generally make a good contribution to promoting knowledge transfer in ZSL. From the fifth row, we can observe performance increase for the CUB benchmark when the entropy loss is removed. This is because that CUB is a fine-grained image recognition benchmark, in which the classes have marginal visual differences. Moreover, the target domain of the CUB benchmark has a relatively large label space. Hence, the unsupervised entropy loss may encourage incorrect predictions for the target domain of CUB. The performance improvement  of the entropy loss is also limited for the SUN benchmark which involves many classes.
	
	In addition, as shown the last four rows of Table~\ref{ablation}, we compare our TSVN with widely used distribution shift alignment approaches, including DANN~\cite{tzeng2017adversarial}, MMD~\cite{long2015learning}, CMD~\cite{zellinger2019robust}, and EMD\cite{shen2018wasserstein}. We can see that the performance is not good if domain adversarial training or MMD is used for distribution alignment. This observation is consistent with our motivation.

	\begin{table}[]
		\renewcommand\arraystretch{1.3}
		\fontsize{9.0}{10.2} \selectfont
		\caption{Comparison for the $\mathcal{A}$-distance between semantic-visual fusions of the source and target domains. For the compared methods, the semantic-visual fusions are constructed by concatenating visual features and semantic attributes.}
		\vspace{0.2cm}
		\centering
		\setlength{\tabcolsep}{3.5mm}{
			\begin{tabular}{crrrr} 
				\hline
				& \multicolumn{1}{c}{CUB} & \multicolumn{1}{c}{aPY} & \multicolumn{1}{c}{AwA} & \multicolumn{1}{c}{SUN}  \\ 
				\hline
				SAE    & 1.31                    & 0.70                    & 1.56                    & 1.23                     \\
				VSOP   & 0.98                    & 1.57                    & 1.68                    & 0.87                     \\
				LisGAN & 1.12                    & 1.78                    & 1.82                    & 1.18                     \\
				\hline
				
				TSVN   & 0.97                    & 0.65                    & 0.74                    & 0.45                     \\
				\hline
		\end{tabular}}
		\label{a-distance}
	\end{table}

	Fig.~\ref{visual} displays the t-SNE visualization for the metric network's intermediate layers. We can see that there exists clear domain discrepancy between the distributions of the semantic-visual fusions from two domains. By using two BN modules to separately normalize the mini-batches from different domains, our design can effectively align the distribution discrepancy. 
	
	Besides the t-SNE visualization, we also quantitatively measure the $\mathcal{A}$-distance \cite{ben2010theory} between semantic-visual fusions of the source and target domains. As shown in Table~\ref{a-distance}, it is clear that our approach achieves the closest $\mathcal{A}$-distance, which verifies the advantage of TSVN in aligning the distribution shift across the source and target domains. This advantage mainly comes from the direct alignment for distribution shift. In contrast, the existing methods can only indirectly bridge the projection shift.

\begin{figure}[t]
		\centering
		{\includegraphics[width=0.36\textwidth]{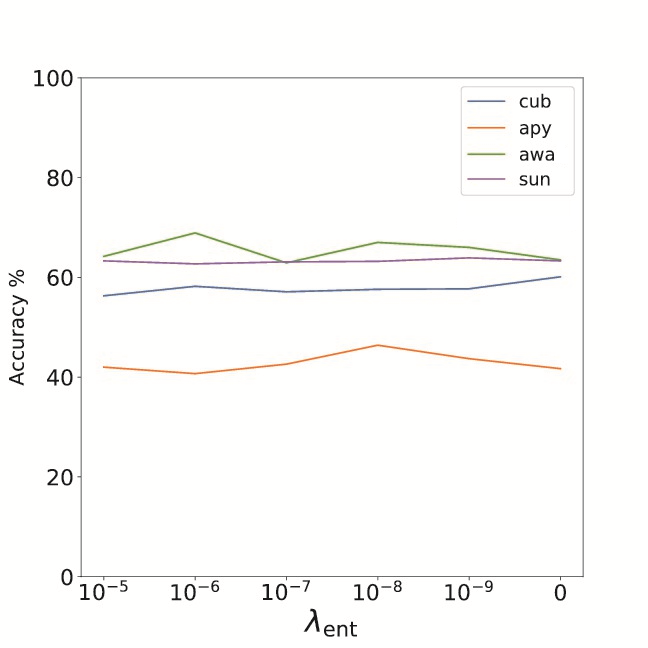}}
		\hfil
		{\includegraphics[width=0.36\textwidth]{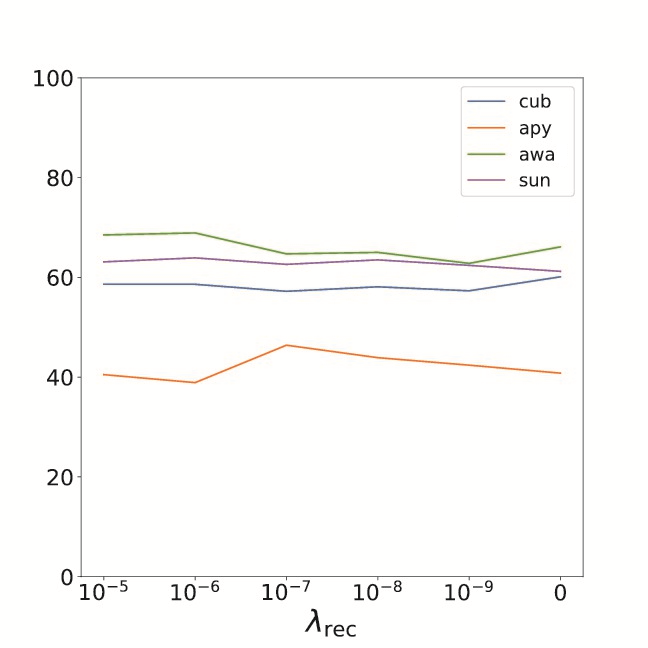}}
		\caption{Sensitivity analysis for the hyper-parameters.}
		\label{sensi}
	\end{figure}

	Finally, we conduct the sensitivity analysis for hyper-parameters  to demonstrate the robustness of our model,  which is clearly displayed in Fig. \ref{sensi}. In particular, we conduct experiments varying the trade-off parameters, including $\lambda_{\rm rec}$ for the attribute reconstruction loss and $\lambda_{\rm ent}$ for the entropy regularization loss. From Fig. \ref{sensi}, we can see that the performance of TSVN is not sensitive to the values of the trade-off parameters.
	
	\section{Conclusion}
	\label{conclusion}
	In this paper, we propose a novel approach dubbed Transferrable Semantic-Visual Network for transductive ZSL. To be specific, we propose to transform transductive ZSL into an unsupervised domain adaptation problem through redrawing ZSL as predicting similarity score for the fusions of semantic attributes and visual features.  To reduce the distribution discrepancy of semantic-visual fusions for the seen and unseen categories, we propose to incorporate two batch normalization units at each BN layer to normalize the mini-batches from the source and target domains separately. As a result, the source and target data can follow similar distribution at each layer of the metric network. Compared with the previous ZSL works which can only indirectly bridge the projection shift, our approach provides a more direct path to bridge the distribution shift across the source and target domains. 
	
	Currently, there still exists a gap between the implementation of our approach and the derived theory. According to the derived theory, the semantic-visual fusions are computed via the tensor product between visual features and semantic attributes. However, 
	the tensor product implementation need to handle high-dimensional semantic-visual fusions which are hard to learn. Hence, we use concatenation to compute the semantic-visual fusions as a compromise. Our future work will further investigate a more efficient semantic-visual fusion manner that can be more close to the derived theory.

	\section*{Acknowledgements}
	This paper is supported by the National Natural Science Foundation of China [grant numbers 11829101, 11931014], and the Fundamental Research Funds for the Central Universities of China [grant number JBK1806002]. The authors would like to thank the anonymous reviewers for the careful reading of this paper and the constructive comments they provided.
	
	
	\bibliographystyle{elsarticle-num} 
	\bibliography{./reference}
	
	
\end{document}